\pdfoutput=1
\documentclass[11pt,a4paper]{article}
\usepackage[hyperref]{acl2020}
\usepackage{times}
\usepackage{latexsym}
\usepackage{graphics}

\usepackage{cjhebrew}
\usepackage{soul, color}

\usepackage{microtype}

\aclfinalcopy 

\DeclareRobustCommand{\hlpink}[1]{{\sethlcolor{pink}\hl{#1}}}

\newcommand{\rt}[1]{\hlpink{ RT: #1}}

\title{{\it AlephBERT:} A  Hebrew Large Pre-Trained Language Model \\to Start-off your Hebrew NLP Application With}

\author{Amit Seker, Elron Bandel, Dan Bareket,  Idan Brusilovsky, Refael Shaked Greenfeld, Reut Tsarfaty\\
Bar-Ilan University, Computer Science Department, Ramat-Gan, Israel\\\{aseker00,elronbandel,dbareket,shakedgreenfeld,brusli1,reut.tsarfaty\}@gmail.com}

\date{}

\begin{document}
\maketitle
\begin{abstract}
Large Pre-trained Language Models (PLMs) have become ubiquitous in the development of language understanding technology and 
lie at the heart of many artificial intelligence advances.
While advances reported for English using PLMs are unprecedented, reported advances using PLMs  in Hebrew are few and far between. The problem is twofold. First, Hebrew resources available for training NLP models are not at the same order of magnitude as their English counterparts.
Second, there are no accepted tasks and benchmarks to evaluate the progress of Hebrew PLMs on. In this work we aim to remedy both aspects. First, we present {\em AlephBERT}, a  large pre-trained language model 
for Modern Hebrew, which is trained on larger vocabulary and a larger dataset than any Hebrew PLM before. Second, using {\em AlephBERT} we present new state-of-the-art results on multiple Hebrew tasks and benchmarks, including: Segmentation, Part-of-Speech Tagging, full Morphological Tagging, Named-Entity Recognition and Sentiment Analysis. 
We make our {\em AlephBERT} model publicly available, providing a single  point of entry for the development of Hebrew NLP applications.
\end{abstract}
\section{Introduction}
Contextualized word representations, provided by models such as BERT \cite{bert} and  RoBERTa \cite{roberta}, were shown in recent years to be critical for obtaining  state-of-the-art performance on a wide range of Natural Language Processing (NLP) tasks --- such as syntactic and semantic parsing, question answering, natural language inference, text summarization, natural language generation, and  more. These  contextualized word representations are obtained by  pre-training a large language model on massive quantities of unlabeled data,  aiming to maximize  a  simple yet effective objective of {masked word prediction}. 

While advances reported for English using such models are unprecedented, in Hebrew previously reported results using BERT-based models are far from impressive. Specifically, the BERT-based Hebrew section of multilingual-BERT  \cite{bert} (henceforth, mBERT), did not provide a similar boost in performance to what is observed for the English section of mBERT. In fact, for several reported tasks, the mBERT model results are on a par with pre-neural models, or neural models based on non-contextialized embeddings \cite{tsarfaty2020,klein2020}. An additional Hebrew BERT-based model, HeBERT \cite{hebert}, has been released,  yet there is no reported evidence on  performance improvements on key component of the Hebrew NLP pipeline --- which includes, at the very least: morphological segmentation,  full morphological tagging, and full (token/morpheme-based) named entity recognition. 

In this work we present {\em AlephBERT}, a  Hebrew  pre-trained language model, larger and  more effective than any Hebrew PLM before. Using {\em AlephBERT} we show  substantial improvements on {all} essential tasks in the Hebrew NLP pipeline, tasks tailored to fit a {\em morphologically-rich language}, including:  {\bf Segmentation, Part-of-Speech Tagging, full morphological tagging, Named Entity Recognition} and {\bf  Sentiment Analysis}. Since previous Hebrew NLP studies used varied  corpora and annotation schemes, we confirm our results on {\em all} existing Hebrew benchmarks and variants. For morphology and POS tagging, we test on {both} the Hebrew section of the SPMRL shared task \cite{spmrl}, and the Hebrew  UD corpus \cite{hebUD}.  For Named Entity recognition, we test on  both the corpus of \citet{bmcner} and that of \citet{nemo}. For sentiment analysis we test on the facebook corpus of \citet{amram2018}, as well as a newer (fixed) variant of this benchmark.

We make 
our pre-trained 
model
publicly available\footnote{\url{huggingface.co/onlplab/alephbert-base}} and additionally we deliver an online demo\footnote{\url{nlp.biu.ac.il/~elronbandel/alephbert/}} allowing to qualitatively compare the mask-prediction capacity of different   PLMs  available for Hebrew. 
In the near future we will release the complete {\em AlephBERT}-geared pipeline we developed, containing the aforementioned tasks, as means for evaluating and comparing future Hebrew PLMs, and as a starting point for developing further downstream applications and tasks. 
We also plan to showcase {\em AlephBERT}'s capacities on downstream language understanding  tasks such as:  Information Extraction, Text Summarization, Reading Comprehension, and more.
As future research, we are pursuing a plan to investigate the effect of different word decomposition algorithms and input representation variants on the different tasks in the Pipeline.
\section{The Challenge}
This paper presents a case study for PLM development for a {\em morphologically-rich} and {\em resource-poor} language. Specifically, we address Modern Hebrew, a Semitic, morphologically-rich  language, that is long known to be notoriously hard to parse.

The challenges posed to automatically processing Hebrew texts and obtaining good accuracies on downstream tasks stem from (at least) two main factors. 
The first is the internal-complexity of word-tokens, resulting from the rich morphology, complex orthography, and lack of diacritization in Hebrew written texts. Space-delimited tokens have non-transparent decomposition and are highly ambiguous, making even the simplest of the tasks in the pipeline very challenging \cite{whatswrong}. 
The second factor is the fact that Modern Hebrew, with only a few dozens of millions of native speakers, is often studied in resource-scarce settings. 

The resource-scarce setting is problematic for PLM development in at least two ways. First, there are insufficient amounts of free unlabeled text for pre-training. To wit, the Hebrew Wikipedia that was the source for training multilingual BERT is of orders of magnitude smaller than the English Wikipedia (See Table \ref{tab:corpora-size-comparison} ).\footnote{Of course, ample Hebrew data does exist online, but most of it is closed due to copy-right issues and paywalls.}  Secondly, there are no large-scale open-access   commonly accepted benchmarks for fine-tuning and/or evaluating the performance of Hebrew PLMs on NLP/NLU downstream tasks. 

Previous studies on various tasks on Hebrew data do exist, each relying on disparate data sources, with varied evaluation metrics and annotation schemes even for the same task.
To investigate Hebrew PLMs and probe their ability to capture linguistic structure, 
we introduce and evaluate Hebrew PLMs on the full set of tasks, sentence-based ,
token-based and morpheme-based tasks, including specific task variants and evaluation metrics.


\begin{table}[t]
\centering
\scalebox{0.9}{
\begin{tabular}{|l||rr|}
\hline
    Language  & Oscar Size & Wikipedia Articles 
    \\\hline  \hline
  English & 2.3T & 6,282,774 \\
  \hline
  Russian & 1.2T & 1,713,164 \\
    \hline
  Chinese & 508G & 1,188,715 \\
  \hline
  French & 282G & 2,316,002 \\
  \hline
  Arabic & 82G & 1,109,879 \\
  \hline
  {\bf Hebrew} & {\bf 20G } & {\bf 292,201} \\
  \hline
\end{tabular}
 }
\caption{
Corpora Size Comparison: High-resource (and Medium-resourced) languages vs.\ Hebrew.
  }\label{tab:corpora-size-comparison}  
  \vspace{-0.05in}
\end{table}

\begin{table}[t]
\centering
\scalebox{0.9}{
\begin{tabular}{|l||rrr|}
\hline
    Corpus  & File Size & Sentences & Words 
    \\\hline  \hline
  Oscar (deduped)   & 9.8GB & 20.9M & 1,043M  \\
  \hline
  Twitter & 6.9GB & 71.5M & 774M  \\
    \hline
  Wikipedia & 1.1GB & 6.3M & 127M \\
  \hline
  {\bf Total} & {\bf 17.9GB} & {\bf 98.7M} & {\bf 1.9B} \\
  \hline
\end{tabular}
 }
\caption{
Data Statistics for AlephBERT's training sets.
  }\label{tab:data-stats} 
  \vspace{-0.05in}
\end{table}

\section{The Model}

\paragraph{Data.}
The PLM nicknamed {\em AlephBERT} is trained on a larger dataset and a larger vocabulary than any Hebrew BERT instantiation before. Data statistics are provided in Table~\ref{tab:data-stats}.
 Specifically, we employ the following datasets for pre-training:

\begin{itemize}
    \item {\bf Oscar:}  A deduplicated Hebrew portion of the OSCAR\- corpus, which is ``extracted from Common Crawl via language classification, filtering and cleaning'' \cite{ortiz-suarez-etal-2020-monolingual}.
    \item {\bf Twitter:} Texts of Hebrew tweets collected between 2014-09-28 and 2018-03-07.  We slightly cleaned up the texts by removing  retweet signals ``{\em RT:}'', user mentions (e.g. ``{\em @username}''), and URLs.
    \item {\bf Wikipedia:} The texts in all of  Hebrew Wikipedia,\footnote{Dump: hewiki-20200201-pages-articles.xml.bz2} extracted using \citet{Wikiextractor2015} This corpus is available on our github.\footnote{\url{https://github.com/OnlpLab/AlephBERT/blob/main/data/wikipedia/}}
\end{itemize}
One of the most important factors driving the success of PLMs in other languages is the availability of enormous amounts of text to learn from.
The Hebrew portions of Oscar and Wikipedia provides us with a training set size which is an order of magnitude smaller compared with  resource-savvy languages, 
as shown in Table~\ref{tab:corpora-size-comparison}.
In order to build a strong PLM we need a considerable boost in the amount of text that the PLM can learn from, which in our case comes form massive amounts of tweets added to the training set.
The textual utterances provided by the Twitter sample API tend be short and diverge from valid syntax and canonical language use for the most part.
And while the free form language expressed in tweets might differ significantly from the text found in Oscar and Wikipedia, the sheer volume of tweets helps us close the resource gap substantially.
Combining all resources together we have tweets comprising the lion's share of sentences in our dataset (72\%).

\paragraph{Training}
We used the Transformers training framework of Huggingface \cite{wolf-etal-2020-transformers} and trained two different models --- a small model with 6 hidden layers learned from the Oscar portion of our dataset, and a base model with 12 hidden layers which was trained on the entire dataset.
The processing units used in both the small and base AlephBERT models are wordpieces generated by training BERT tokenizers over the respective datasets with a vocabulary size of 52K in both cases.

Traditionally, BERT models are optimized with an objective function optimized using both masked token prediction as well as next sentence prediction losses.
Following the work on RoBERTa \cite{roberta} we employ masked-token prediction loss only in our training objective. 
Incidentally our choice of dataset also forces us to ignore next sentence prediction because a large portion of our data comprises of tweets  which are unrelated and independent of each other (we did not attempt to reconstruct the discourse threads  of retweets and replies).
For more training details see the Appendix.

\section{Experiments}
\paragraph{Goal}
We set out to pre-train Hebrew PLMs and evaluate them empirically on a range of Hebrew NLP tasks. 
We evaluated the two {AlephBERT} variants (small and base)   on the different  tasks, in order to empirically gauge the effect of model size and data size on the quality of the language model.
In addition, we compared the performance of our models to existing Hebrew BERT-based instantiations (mBERT \cite{bert} and HeBERT \cite{hebert}). 
We evaluated the PLMs  on all key tasks of the Hebrew NLP pipeline.

\paragraph{Benchmarks} 
We evaluate our BERT-based models on various Hebrew NLP tasks   using the following   benchmarks:

\begin{itemize}
    \item {\bf Word Segmentation, Part-of-Speech Tagging, Full Morphological Tagging:}
    \begin{itemize}
        \item The Hebrew Section of the SPMRL Task \cite{spmrl}
        \item The Hebrew Section of the UD\footnote{\url{https://universaldependencies.org}} treebanks collection \cite{hebUD}
    \end{itemize}
    \item {\bf Named Entity Recognition:}
    \begin{itemize}
        \item Token-based NER evaluation based on the corpus of Ben-Mordecai and Elhadad \cite{bmcner}
        \item Token-based and Morpheme-based NER evaluation based on the Named Entities and MOrphology (henceforth NEMO) corpus  \cite{nemo}
    \end{itemize}
    \item {\bf Sentiment Analysis:}
   \begin{itemize}
       \item 
    Sentiment Analysis  evaluation based on the corpus of  \citet{amram2018}.
    \item Since the aforementioned corpus is reported to be leaking (shared material between test and train), we provide a cleaned up version and evaluate on the updated split.
    \end{itemize}
\end{itemize}

\section{Tasks and Modeling Strategies}
A key question when assessing BERT-based PLM performance for Hebrew  concerns how to develop models for the different levels of granularity. Here we briefly sketch our modeling strategies, starting with the easiest (classification) tasks and continuing to the more challenging setups, involving the use of PLMs to predict  the tokens' internal  structures.
\subsection{Sentence-Based Modeling}

\paragraph{Sentiment Analysis}
The first task we report on is a simple sentence classification task, classifying the sentiment of a given sentence to one of three values: negative, positive, neutral.
We trained and evaluated  BERT-based sentence classification on two variants of the Hebrew Sentiment dataset of \citet{amram2018}. 

The first variant is the original sentiment dataset of \citet{amram2018} with an additional split to create a dev set (the original paper had only train and test split, and the test set remains the same). The dev set contains 10\% of the train data  which leaves us with a split of 70-10-20. 

Unfortunately, the original dataset of Amram et al.\ had  a significant data leakage between the splits, with  duplicates in the data samples. After removing the duplicates out of the original 12,804 sentences, we are left with a dataset of size 8,465.\footnote{\url{https://github.com/OnlpLab/Hebrew-Sentiment-Data}} 

We fine-tuned all the models for 15 epochs with the default Huggingface \cite{wolf-etal-2020-transformers} parameters on 5 different seeds. We report per-comment accuracy, and take the mean of these 5 runs.


\subsection{Token-Based  Modeling}

\paragraph{Named Entity Recognition}
For the NER task, we initially assume a token-based sequence labeling model. The input comprises of the sequence of tokens in the sentence, and the output contains  BIOES tags indicating entity spans. 
The token-based model is a simple fine-tuned model using the Transformer's token-classification script of \citet{wolf-etal-2020-transformers}.

We evaluate this model on two corpora. The first is the corpus by \citet{bmcner}, henceforth, the BMC corpus. 
The BMC corpus annotates entities at Token-level. This means that a Hebrew token containing both a preposition and an entity mention will not deliver the entity-mention boundaries. The BMC contains 3294 sentences and 4600 entities, and has seven different entity categories (DATE, LOC, MONEY, ORG, PER, PERCENT, TIME). To remain compatible with the original work we train and test the models on the 3 different splits as in \citet{nemo}.\footnote{\url{https://github.com/OnlpLab/HebrewResources/tree/master/BMCNER}} 
For the BMC corpus we report token-based  F1 scores on the detected entity mentions.

The second corpus is an extension of the SPMRL dataset with Named Entities annotation, also marked by BIOSE tags, respecting the precise (token-internal) morphological boundaries of NEs  (henceforth, NEMO, standing for Named Entities and MOrphology) \cite{nemo}. This corpus provides both a token-based and a morpheme-based annotation of the entities, where the latter contains the accurate (token-internal) entity boundaries. The NEMO corpus has nine categories (ANG, DUC, EVE, FAC, GPE, LOC, ORG, PER, WOA).   It contains 6220 sentences and 7713 entities, and  we used the standard SPMRL Train-Dev-Test, as in \citet{nemo}

The models were trained over 15 epochs and no hyper parameter tuning. For the BMC we used 3 different seeds for each split set, leading  to overall nine different training rounds, and for the NEMO set we used the average mean of five different seeds.  
For both benchmarks we report token-based F1 scores on the detected entity mentions.

\subsection{Morpheme-Based Modeling}
Modern Hebrew is a Semitic language with rich morphology and complex orthography. As a result, the basic processing units in the language are typically smaller than a given token's span. To probe {AlephBERT}'s capacity to accurately predict such token-internal  linguistic structure,  we test our  models on four  tasks that require knowledge of the internal morphology  of the raw tokens:

\begin{itemize}
\item  {\bf Segmentation}\\
\underline{Input:} A Hebrew sentence containing raw space-delimited tokens \\
\underline{Output:} A sequence of morphological segments representing  basic processing units.\footnote{These units comply with the 2-level representation of tokens defined by UD, where each basic unit corresponds to a single POS tag. \url{https://universaldependencies.org/u/overview/tokenization.html}}
\item  {\bf Part-of-Speech Tagging}\\
\underline{Input:} A Hebrew sentence containing raw  space-delimited tokens \\
\underline{Output:} Segmentation of the tokens to basic processing units as above, where each segment is tagged with its single disambiguated part-of-speech tag.
\item   {\bf Morphological Tagging}\\
\underline{Input:} A Hebrew sentence containing raw space-delimited tokens \\
\underline{Output:} Segmentation of the tokens to basic processing units as above, where each segment is tagged with a single POS tag and a set of morphological features.\footnote{Equivalent to the AllTags evaluation metric defined in the CoNLL18 shared task. \url{https://universaldependencies.org/conll18/results-alltags.html}}
\item   {\bf Morpheme-Based NER}\\
\underline{Input:} A Hebrew sentence containing raw space-delimited tokens \\
\underline{Output:} Segmentation of the tokens to basic processing as above where segment is tagged with a BIOSE tags indicating entity spans, along with the entity-type label.
\end{itemize}
An illustration of these tasks is given in Table \ref{tab:tasks}.

\begin{table*}[t]
\scalebox{0.88}{
\begin{tabular}{|l||c|c|c|c|c|}
\hline
Raw input & \multicolumn{5}{c|}{\cjRL{lbyt hlbn}} \\ 
\hline
Space-delimited tokens & \multicolumn{2}{c|}{\cjRL{hlbn}} & \multicolumn{3}{c|}{\cjRL{lbyt}} \\
\hline
\hline
Segmentation & \cjRL{lbn} & \cjRL{h} & \cjRL{byt} & \cjRL{h} & \cjRL{l}\\
\hline
POS &  ADJ & DET & NOUN & DET & ADP\\
\hline
Morphology &  \small{Gender=Masc$\vert$Number=Sing} & \small{PronType=Art} & \small{Gender=Masc$\vert$Number=Sing} & \small{PronType=Art }& \small{-} \\
\hline
Token-level NER  & \multicolumn{2}{c|}{E-ORG} & \multicolumn{3}{c|}{B-ORG}  \\
\hline
Morpheme-level NER &  E-ORG & I-ORG & I-ORG & B-ORG & O \\
\hline
\end{tabular}
}
\caption{\label{tab:tasks} 
Illustration of Evaluated Token and Morpheme-Based Downstream Tasks. The input is the two-word input phrase ``\cjRL{lbyt hlbn}'' ({\em to the White House}). Sequence and Hebrew text goes from right to left.
  }
\end{table*}

As opposed to fine-tuning the PLM model parameters, as done in sentence-based and token-based classification tasks, segmented morphemes are not readily available in the BERT representation. 
In order to provide proper segmentation and labeling for the four aforementioned tasks we developed a model designated to produce the morphological segments of  each token in context.

The morphological segmentation model which we designed is composed of a PLM responsible for transforming input tokens into contextualized embedded vectors, which we then feed into a char-based seq2seq module that extracts the output segments.
The seq2seq module is composed of an encoder implemented as a simple char-based BiLSTM, and a decoder implemented as a char-based LSTM generating the output character symbols, or a space symbol signalling the end of a morphological segment. 
We train the model for 15 epochs, optimizing next-character prediction loss function.

For the other tasks, involving both segmentation and labeling we deploy an MTL (multi-task learning) setup. That is, when generating an end-of-segment symbol, the model then predicts task labels which can be one or more of the following: POS-tag, NER-tag, morphological features. In order to guide the training to learn we optimize the combined segmentation and label prediction loss values. Currently we simply add together the loss values, but we note that as a future improvement it is likely that assigning different weights to the different loss values could prove to be beneficial.
All of these morphological labeling models are trained for 15 epochs and evaluated on both the UD \cite{hebUD} and SPMRL data \cite{spmrl}.

In addition, we design another setup for running the various morphological labeling tasks in which we first segment the text (using the above-mentioned segmentation model) and then perform fine-tuning with a token classification attention head directly applied to the PLM (similar to the way we fine-tune the PLM for the token-based NER task described in the previous section).
In this pipeline setup we utilize the PLM twice; as part of the segmentation model to generate segments, which we then feed directly into the PLM (augmented with a token classification head) which is fine-tuned for the specific labeling task.
We acknowledge the fact that we are fine-tuning the PLM using morphological segments even though it was originally pre-trained without any knowledge of sub-token units. But, as we shall see shortly, this seemingly unintuitive strategy performs surprisingly well.

\section{Results}
\paragraph{Sentence-Based Tasks}
The Sentiment analysis experimental results are provided in Table \ref{sentiment-results}. As can be seen, all BERT-based models substantially outperform the original CNN Baseline reported by \citet{amram2018}. Interestingly, both AlephBERT-small and AlephBERT-base outperform all BERT-based variants, with BERT-base setting new SOTA results on the new (fixed) dataset.

\paragraph{Token-Based Tasks}
For our two NER benchmarks, we report the NER F1 scores  on the token-based fine-tuned model  in Table \ref{ner-results}.

Here, although we see noticeable improvements for the mBERT and HeBert variants over the current SOTA, the most significant increase is in the AlephBERT-base model. We also see a substantial difference between the AlephBERT-small and AlephBERT-base models, with the latter providing a new SOTA results on these both data sets. Crucially, this holds for the {\em token-based} evaluation metrics  (as defined in \citet{nemo}).
\begin{table}[t]

\begin{tabular}{|r||cc|}
\hline
       & NEMO  & BMC
   \\\hline   \hline
  Previous SOTA     & 77.75   &  85.22  \\
  \hline
  mBERT  & 79.07  & 87.77 \\
  HeBERT &  81.48  &  89.41 \\
    \hline
  AlephBERT-small &  78.69  &  89.07 \\
    AlephBERT-base & \textbf{84.91}   & \textbf{91.12} \\
  \hline
\end{tabular}
\caption{
Token-Based NER Results on the NEMO and the Ben-Mordecai Corpora. Previous SOTA on both corpora has been reported by the NEMO models of \citet{nemo}. }\label{ner-results}
\end{table}
 

 \begin{table*}
\begin{tabular}{|r||cc|cc|}
\hline
    &  Old(leak) token & Old(leak) morph & New(fixed) token & New(fixed) morph
   \\\hline   \hline
  Previous SOTA   & 89.2 & 87.5 &NA & NA\\
  \hline
  mBERT & 92.12 & 92.18 & 84.21 & 85.58 \\
  HeBERT & 92.48 & 92.27 & 87.13 & 86.88 \\
    \hline
  AlephBERT-small & \textbf{93.15} & \textbf{92.70} & 88.3 & 87.38 \\
    AlephBERT-base& 91.63 & 92.01  & \textbf{89.02} & \textbf{88.71}\\
  \hline
\end{tabular}
\caption{Sentiment Analysis Scores on the Facebook Corpus. Previous SOTA is  reported by \citet{amram2018}.}
\label{sentiment-results}
\end{table*}

\begin{table*}
\centering

\begin{tabular}{|r||ccc|}
\hline
    & Segmentation F1 & POS F1 & Morphological Features F1
    \\\hline  \hline
  Previous SOTA   & NA & 90.49 & 85.98 \\
  \hline
  mBERT-morph & 97.36 & 93.37 & 89.36 \\
  HeBERT-morph & 97.97 & 94.61 & 90.93 \\
    \hline
  AlephBERT-small-morph & 97.71 & 94.11 & 90.56 \\
  AlephBERT-base-morph & {\bf 98.10} & {\bf 94.90} & {\bf 91.41} \\
  \hline
\end{tabular}
\caption{Morpheme-Based Aligned MultiSet (mset) Results on the SPMRL Corpus. Previous SOTA is as reported by \cite{seker-tsarfaty-2020-pointer} (POS) and \cite{more2019tacl} (morphological features)
}\label{morph-results-spmrl-mset}
\end{table*}
\begin{table*}

\begin{tabular}{|r||cccc|}
\hline
    & Segmentation F1 & POS F1 & Morphological Features F1 &  
   \\\hline  \hline
  Previous SOTA   & NA & 94.02 & NA &   \\
    \hline
  mBERT-morph & 97.70 & 94.76 & 90.98 &   \\
  HeBERT-morph & 98.05 & 96.07 & 92.53 &   \\
    \hline
  AlephBERT-small-morph & 97.86 & 95.58 & 92.06 &   \\
  AlephBERT-base-morph & {\bf 98.20} & {\bf 96.20} & {\bf 93.05} &   \\
  \hline
\end{tabular}
\caption{Morpheme-Based Aligned MultiSet (mset) Results on the UD
Corpus. Previous SOTA is as reprted by \cite{seker-tsarfaty-2020-pointer} (POS)}\label{morph-results-ud-mset}
\end{table*}

\begin{table*}

\begin{tabular}{|r||cccc|}
\hline
    & Segmentation F1 & POS F1 & Morphological Features F1 & 
   \\\hline  \hline
  Previous SOTA   & 96.03 & 93.75 & 91.24 & \\
    \hline
  mBERT-morph & 97.17 & 94.27 & 90.51 &   \\
  HeBERT-morph & 97.54 & 95.60 & 92.15 &   \\
    \hline
  AlephBERT-small-morph & 97.31 & 95.13 & 91.65 &   \\
  AlephBERT-base-morph & {\bf 97.70} & {\bf 95.84} & {\bf 92.71} &   \\
  \hline
\end{tabular}

\caption{Morpheme-Based Aligned (CoNLL shared task) Results on the UD Corpus. Previous SOTA is as reported by \citet{nguyen2021trankit}}\label{morph-results-ud-conll}
\end{table*}

\begin{table*}[t]
    \centering
    \begin{tabular}{| *{9}{c|} }
    \hline
    Architecture & \multicolumn{2}{c|}{Pipeline}
            & \multicolumn{2}{c|}{Pipeline}
                    & \multicolumn{2}{c|}{MultiTask} \\
    Segmentation & \multicolumn{2}{c|}{(Oracle)}
            & \multicolumn{2}{c|}{(Predicted)}
                    & \multicolumn{2}{c|}{} \\
    Scores (aligned mset F1) & Seg & NER & Seg & NER & Seg & NER \\
    \hline\hline
    Previous SOTA (NEMO) & 100.00 & 79.10     &  95.15   &   69.52  &   97.05    &  77.11    \\
    \hline
    mBERT          &100.00 & 77.92 & 97.68 & 72.72 & 97.24 & 72.97 \\
    HeBERT         & 100.00 & 82    & 98.15 & 76.74 & 97.92 & 74.86 \\
    \hline\hline
    AlephBERT-small& 100.00 & 79.44 & 97.78 & 73.08 & 97.74 & 72.46 \\
    AlephBERT-base   & 100.00 & 83.94 & {\bf 98.29} & {\bf 80.15} & 98.19 & 79.15 \\
    \hline
    \end{tabular}
    \caption{Morpheme-Based NER Evaluation on the NEMO Corpus. Previous SOTA is as reported by \citet{nemo} for the Pipeline (Oracle), Pipeline (Predicted) and a Hybrid (almost-joint) Scenarios, respectively.
    }\label{ner-morph}
    \label{morph-ner-eval}
\end{table*}

\paragraph{Morpheme-Based Tasks}

As a particular novelty of this work, we  report BERT-based results on sub-token (segment-level) information. Specifically, we evaluate segmentation F1, POS F1, Morphological Features F1 and morphem-base NER F1, compared against the disambiguated labeled segments.
In all cases we use raw space-delimited tokens as input, letting the BERT-based models perform {\em both} the segmentation and labeling. 

Table \ref{morph-results-spmrl-mset} 
presents the segmentation, POS tags, and morphological tags F1  for the SPMRL dataset, all evaluated at the granularity of morphological segments.
We report the aligned multiset F1 Scores as in previous work on Hebrew    \cite{more2019tacl}.

We see that segmentation results for all BERT-based models are  similar, and they are already at the higher range of 97-98 F1 scores, which are hard to improve further.\footnote{Some of these errors are due to annotation errors, or truly ambiguous cases.}  For POS tagging and morphological features, all BERT-based models significantly outperform the previous SOTA provided by \cite{seker-tsarfaty-2020-pointer} (referred to as PtrNet) for POS tags and \cite{more2019tacl} (referred to as YAP) for morphological features. 
With respect to all BERT-based variants, we see an improvement for AlephBERT on all other alternatives, but on a  small scale. That said, we do notice a repeating trend that places   AlephBERT-base as the best model for all of our morphological tasks, indicating that the improvement provided by the depth of the model and  a larger dataset does also improve the ability to capture token-internal structure.

These trends are replicated on the UD Hebrew corpus, for two different evaluation metrics --- the Aligned MultiSet F1 Scores as in previous work on Hebrew    \cite{more2019tacl}, \cite{seker-tsarfaty-2020-pointer},
and the Aligned F1 scores  metrics in the UD shared task \cite{zeman-ud} --- as reported in Tables \ref{morph-results-ud-mset} and \ref{morph-results-ud-conll} respectively. AlephBERT obtains the best results for all tasks, even if not by a large margin.


\paragraph{Morpheme-Based NER}
Earlier in this section we considered NER as a token-based task that simply requires fine-tuning on the token labels. However, this setup is not accurate enough and less useful for downstream tasks, since the exact entity boundaries are often token internal \cite{nemo}.
We hence also report here morpheme-based NER evaluation, respecting the exact  boundaries of the Entity mentions.
To obtain morpheme-based labeled-span of Named Entities as discussed above we could either employ a pipeline, first predicting segmentation and then applying a fine tuned labeling model {\em directly on the segments}, 
or we can use the MTL model and predict NER labels  {\em while}  performing the segmentation.

Table \ref{ner-morph} presents segmentation and NER results for three different scenarios: (i) a pipeline assuming 
gold segmentation (ii)  a pipeline assuming the best predicted segmentation (as predicted above) (iii) obtaining the segmentation and NER labels jointly in the MTL setup.

As our results indicate, AlephBERT-base consistently scores highest in both pipeline (oracle and predicted) and multi-task setups.
Looking at the Pipeline-Predicted scores, there is a clear correlation between a higher segmentation quality of a PLM and its ability to produce better NER results.
Moreover, the differences in NER scores between the models are considerable (unlike the subtle differences in segmentation, POS and morphological features scores) and draw our attention to the relationship between the size of the PLM, the size of the pre-training data and the quality of the final NER models.
Specifically, HeBERT and AlephBERT-small were pre-trained with similar datasets - HeBERT with Oscar and Wikipedia, AlephBERT-small with Oscar only (the Wikipedia portion is order of magnitude smaller compared with Oscar) and comparable vocabulary sizes (heBERT with 30K and AlephBERT-small with 52K). 
However we notice that HeBERT, with its 12 hidden layers,  performs significantly better compared to AlephBERT-small which is composed of only 6 hidden layers. It thus appears that semantic information is learned in those deeper layers which helps in both learning to discriminate entities  and improve the overall morphological segmentation capacity.

In addition, comparing HeBERT to AlephBERT-base we point to the fact that they are both modeled with the same 12 hidden layer architecture, the only differences between them are in the size of their vocabularies (30K vs 52K respectively) and the size of the training data (Oscar-Wikipedia vs Oscar-Wikipedia-Tweets). The improvements exhibited by AlephBERT-base, compared to HeBERT, suggests that it is  a result of the large amounts of training data and larger vocabulary available in our setup. By exposing AlephBERT-base to an amount of text which order of magnitude larger we increased its NER capacity.

Finally, our NER experiments suggest that a pipeline composed of our near-to-perfect morphological segmentation model followed by AlephBERT-base augmented with a token classification head is the best strategy for generating morphologically-aware NER labels.




\section{Qualitative Assessment}
To allow for qualitative assessment of the PLMs, we deliver an online demo where one can compare the {\em masked-word prediction} capacities of the different models, and get the impression of the strengths and weaknesses. 
Our demo, available at \url{https://nlp.biu.ac.il/~elronbandel/alephbert/}, offers friendly graphical interface that allows one to mask an item in a running Hebrew text and obtain the top-N list of alternatives predicted by each of the models. The demo allows to explore the predictions of our models both at token level and at sub-token level, masking individual word-pieces. 
Note that the {AlephBERT} family of models is still under development, and we will add new model variants as we proceed. Stay tuned!

\section{Conclusion}

Modern Hebrew, a morphologically rich and resource-scarce language, has for long suffered from a gap in the resources available for NLP applications, and lower level of empirical results than observed in other,  resource-rich languages. This work provides the first step in remedying the situation, by making available a large Hebrew PLM, nicknamed {AlephBERT}, with larger vocabulary and larger training set than any Hebrew PLM before, and with clear evidence as to its empirical advantages.
Our  {AlephBERT}-base model obtains state-of-the-
art results on the tasks of  segmentation, Part of Speech Tagging, Named Entity Recognition, and Sentiment Analysis. We outperform both general multilingual PLMs (mBERT) as well as language specific instantiations (HeBERT). 
More importantly, using the new AlephBERT models we are now gaining similar benefits as achieved in high resource languages from PLMs.

\section{Acknowledgements}
We are enormously grateful to Roee Aharoni from Google and Yoav Goldberg from Bar Ilan University for technical advise during the project. No less importantly, we are indebted to Roee Aharoni for coining the brand-name {\em AlephBERT}. The research reported in this paper is funded by an individual grant by the Israel Science Foundation (ISF grant \verb|#|1739/26) and a starting grant by the European Research Council (ERC-StG Grant \verb|#|677352), for which we are grateful.

\begin{table*}[t]
    \centering
\begin{tabular}{|r|cccc|}
\hline
    & AlephBERT-base & AlephBERT-small & HeBERT & mBERT-cased \\
   \hline
  max\_position\_embeddings & 512 & 512 & 512 & 512 \\
  num\_attention\_heads & 12 & 12 & 12 & 12 \\
  num\_hidden\_layers & 12 & 6 & 12 & 12 \\
  vocab\_size & 52K & 52K & 30K & 120K\textdagger \\
  \hline
\end{tabular}
\caption{
 Huggingface BERT Configurations Comparison.
{\textdagger Only 2450 vocabulary entries contain Hebrew letters} }\label{train-params}
\end{table*}

\bibliography{acl2020}
\bibliographystyle{acl_natbib} 

\appendix
\section{AlephBERT Training Details}
For reference and to make our work reproducible we specify here the main steps taken and parameters used during training of AlephBERT.
We utilized the Huggingface Transformers framework with most of the default training parameter values. Table-\ref{train-params} lists all of the training parameters that we have manually specified in our code. We also list the values used by the other models.

Training our AlephBERT-base model using the entire dataset proved to be technically challenging due to the model size and data size. With the naive approach training the entire dataset without splitting it into chunks did not utilize the full processing capacity of the GPUs and would have taken several weeks to complete. To overcome this issue we followed the advice to split the dataset into chunks based on the number of tokens in a sentence. The first chunk consisted of 70M senetences with 32 or less tokens. By limiting the maximum number tokens we consequently limit the size of the training matrices used by this chunk which consequently allowed for significantly increasing the batch size which resulted in dramatically shorter training time - these 70M sentences took only 2.5 days to complete 5 epochs. The second chunk consisted of sentences having between 32 and 64 tokens, the third chunk between 64 and 128 and the final last chunk all sentences with more than 128 tokens. 
We trained each chunk for 5 epochs with setting the learning rate to 1e-4. Once we went over the entire dataset we trained for another 5 epochs with a learning rate set to 5e-5 for a total of 10 epochs.
We trained our base model over the entire dataset for 10 epochs on a NVidia DGX server with 8 V100 GPUs which took 8 days.
The small model was trained over 10 epochs using 4 GTX 2080ti GPUs for 5 days in total.

\end{document}